\documentclass{article}

\usepackage[preprint]{neurips_2026}
\makeatletter
\renewcommand{\@noticestring}{}  
\makeatother

\usepackage[utf8]{inputenc}
\usepackage[T1]{fontenc}
\usepackage{hyperref}
\usepackage{url}
\usepackage{booktabs}
\usepackage{amsfonts}
\usepackage{amsmath}
\usepackage{nicefrac}
\usepackage{microtype}
\usepackage{xcolor}
\usepackage{graphicx}
\usepackage{multirow}
\usepackage{array}
\usepackage{tabularx}

\title{Unsolvability Ceiling in Multi-LLM Routing:\\
An Empirical Study of Evaluation Artifacts}

\author{%
  Saloni Garg \\
  Independent Researcher \\
  San Francisco, California, USA \\
  \And
  Amit Sagtani \\
  San Francisco State University \\
  San Francisco, California, USA \\
}

\begin{document}

\maketitle

\begin{abstract}
Efficient routing across multiple LLMs is central to cost-effective deployment, enabling better
trade-offs between cost, latency, and quality. Prior work has attributed substantial routing
headroom to an ``unsolvability ceiling'': a non-trivial fraction of queries that no single model in
the pool can solve correctly. In this paper, we present a large-scale study of multi-tier LLM
routing with over 206,000 query-model pairs across six diverse benchmarks (MMLU, MedQA,
HumanEval, MBPP, Alpaca, ShareGPT) using the Gemma~4 and Llama~3.1 model families.
Evaluating with both LLM-as-a-judge and exact-match metrics, we show that a substantial portion
of reported unsolvability stems from evaluation artifacts rather than fundamental model
limitations. We identify three primary sources: (i)~systematic judge biases favoring specific
response styles and verbosity, (ii)~truncation under fixed generation budgets, and
(iii)~output format mismatches. Through dual-judge validation and exact-match grounding, we
consistently reduce measured unsolvability across tasks. We introduce a decomposition framework
that attributes failures to these artifacts, revealing consistent patterns across domains and model
families. These artifacts also distort supervision signals for routing models. Standard routers
trained on oracle labels are highly sensitive to class imbalance and frequently collapse to the
majority class (${\sim}79\%$ smallest-tier optimal), a behavior confirmed via random-feature and
shuffled-label controls. This leads to a 13-17 percentage point opportunity cost in routing
performance. We provide actionable recommendations, including dual-judge validation, exact-match
anchoring, and cost-sensitive objectives for more robust evaluation and training. Our findings
suggest that existing routing headroom estimates are substantially inflated, calling into question
many reported gains and underscoring the need for reliable protocols in multi-LLM systems.
\end{abstract}

\section{Introduction}
\label{sec:intro}

The problem of routing queries to the most cost-effective language model capable of answering
them correctly has attracted significant research interest. FrugalGPT~\citep{chen2023frugalgpt}
demonstrated that cascading queries across models of increasing capability can reduce cost by up
to 98\% while matching the best individual model's performance. RouteLLM~\citep{ong2024routellm}
introduced a training framework for binary routers using human preference data, achieving over
$2\times$ cost savings on standard benchmarks. A central assumption underlying these gains is that
routing headroom is real and recoverable: a non-trivial fraction of queries are genuinely
unsolvable by the cheapest model but solvable by larger ones, and that this gap can be exploited
by routing systems.

This assumption rests on the quality of the evaluation used to measure it. When routing headroom
is estimated from LLM-as-a-judge scores, the measured headroom reflects not only genuine model
capability differences but also any systematic tendencies of the judge to reward or penalize model
outputs for reasons unrelated to correctness. When it is estimated from fixed-budget inference,
the measured headroom reflects not only model capability but also truncation effects that prevent
models from completing responses. When it is estimated from free-form generation on
multiple-choice tasks, the measured headroom also depends on whether models express their answer
in the expected format.

We present a large-scale empirical study designed to disentangle genuine unsolvability from these
evaluation artifacts. We score 206,756 query-model pairs (51,689 queries $\times$ 4 tiers) from
six diverse benchmarks using the Gemma~4 model family deployed on self-hosted H100 infrastructure
with thinking mode disabled and greedy decoding. We validate judge scores against exact-match
accuracy on two benchmarks with verifiable gold labels (MMLU, MedQA), run cross-family
validation with Llama-3.1-8B and Llama-3.1-70B, and train three router variants to characterize
how evaluation artifacts propagate into routing model training.

\paragraph{Central finding.}
A substantial fraction of reported unsolvability is attributable to three systematic evaluation
artifacts rather than fundamental model limitations.

\paragraph{Artifact~1: Evaluation misalignment.}
Our judge (a state-of-the-art 26B mixture-of-experts model) diverges systematically from
exact-match ground truth on both knowledge benchmarks, but in opposite directions: it assigns
lower scores than exact-match on MMLU by 10-24 pp across all tiers, and higher scores than
exact-match on MedQA for large models by up to 6~pp. The judge partially conflates correctness
with fluency, structure, and reasoning presentation, partially crediting responses with strong
reasoning but incorrect final answers, and partially penalizing responses with correct answers
expressed in terse or unconventional form.

\paragraph{Artifact~2: Truncation.}
Per-source maximum token budgets are binding for MMLU (65\% truncation rate) and MedQA
(57\% truncation rate). Truncated responses frequently receive score~$<2$ from the judge even
when the model has identified the correct answer.

\paragraph{Artifact~3: Output format mismatch.}
Parse failures on MMLU range from 5.4\% to 12.4\% across tiers; on MedQA from 0.6\% to 5.2\%.
Format mismatch preferentially inflates apparent unsolvability for smaller tiers and compresses
the measured quality gradient.

\paragraph{Routing supervision.}
These artifacts compound in routing model training. Oracle labels derived from judge-scored data
reach 79.3\% E2B-optimal by construction, and standard unweighted routers collapse entirely to
majority-class prediction. Random-feature and shuffled-label controls confirm the collapse is
caused entirely by the label marginal. The opportunity cost (routing performance achievable with
dual-judge oracle labels and cost-sensitive objectives versus collapsed routers) is
\textbf{13-17 percentage points} across knowledge-intensive benchmarks.

\paragraph{Contributions.}
(1)~An artifact decomposition framework identifying and quantifying three sources of
evaluation-induced unsolvability inflation; (2)~a dual-judge validation methodology pairing
LLM-as-judge with exact-match grounding; (3)~cross-family ceiling analysis showing the ceiling is
model-family-specific for knowledge tasks and near-universal for conversational tasks;
(4)~a characterization of routing collapse under realistic label distributions with controlled
experiments; and (5)~actionable recommendations for evaluation protocol design.

\section{Model Family and Infrastructure}
\label{sec:models}

\subsection{The Gemma~4 Tier Family}

Table~\ref{tab:tiers} describes the four inference tiers. All models are from the Gemma~4 family
(Google DeepMind), licensed under Apache~2.0, downloaded from HuggingFace on 2026-04-21.

\begin{table}[t]
\caption{Gemma~4 inference tier specifications. The 26B-A4B MoE model activates only ${\sim}4$B
  parameters per forward pass (equal active compute to E4B) while carrying 26B total parameters
  of knowledge capacity.}
\label{tab:tiers}
\centering
\small
\begin{tabular}{llllr}
\toprule
Tier & HuggingFace Model ID & Architecture & Total & Active \\
\midrule
E2B     & \texttt{google/gemma-4-e2b-it}      & Dense, 35L, GQA 8/1       & ${\sim}2$B  & ${\sim}2$B  \\
E4B     & \texttt{google/gemma-4-e4b-it}      & Dense, 42L, GQA 8/2       & ${\sim}4$B  & ${\sim}4$B  \\
26B-A4B & \texttt{google/gemma-4-26b-a4b-it}  & MoE, 30L, 128 experts     & ${\sim}26$B & ${\sim}4$B  \\
31B     & \texttt{google/gemma-4-31b-it}      & Dense, 60L, GQA 32/16     & ${\sim}31$B & ${\sim}31$B \\
\bottomrule
\end{tabular}
\end{table}

E4B and 26B-A4B share approximately equal active parameter counts (${\sim}4$B per forward pass),
predicting near-identical inference latency. The MoE model's 26B total parameters encode
substantially more knowledge than a 4B dense model: equal inference cost, substantially different
knowledge capacity.

\subsection{Inference Infrastructure}

All models are served using vLLM~0.9.1 on dedicated H100~80GB HBM3 GPUs with bf16 precision,
4,096-token context, and \texttt{gpu-memory-utilization=0.90}. Thinking mode is disabled globally
via \texttt{--default-chat-template-kwargs `\{"enable\_thinking": false\}'} and per-call via
\texttt{extra\_body=\{"enable\_thinking": False\}}.

\subsection{Primary Judge Model}

We use \texttt{google/gemma-4-26b-a4b-it}, the same weights as the 26B-A4B inference tier, as
the primary judge, with \texttt{--reasoning-parser gemma4} enabled. The judge applies a 0/1/2
rubric: 0~= incorrect or unhelpful, 1~= partially correct or incomplete, 2~= correct and complete.
Because the judge shares weights with one evaluated tier, we flag this as a limitation and
recommend a held-out judge from a different family in future work (Section~\ref{sec:limits}).

\paragraph{The need for dual-judge validation.}
For benchmarks with verifiable gold labels, relying solely on the LLM judge risks conflating
evaluation artifacts with genuine model failures. Discordance between the two judges reveals cases
where the two evaluators operationalize ``correctness'' differently (the LLM judge through
reasoning quality and completion, exact-match through terminal answer agreement), enabling us to
attribute divergences to specific evaluation properties rather than model failures alone.

\section{Experimental Setup}
\label{sec:setup}

\subsection{Query Dataset}

Table~\ref{tab:dataset} summarizes 51,689 queries drawn from six benchmarks.

\begin{table}[t]
\caption{Dataset composition. MMLU is 9,951 queries after deduplication (49 duplicates removed).
  ShareGPT is 16,318 after filtering to valid human turns.}
\label{tab:dataset}
\centering
\small
\begin{tabular}{lrllr}
\toprule
Source & $N$ & Split & Avg.\ Tokens & Domain \\
\midrule
ShareGPT  & 16,318 & eval only               & 183 & Conversation \\
Alpaca    & 15,000 & 12k train / 3k test     &  27 & Instruction following \\
MedQA     & 10,000 & eval only               & 211 & Medical knowledge \\
MMLU      &  9,951 & eval only               & 115 & Multi-subject knowledge \\
MBPP      &    256 & eval only               &  27 & Code generation \\
HumanEval &    164 & eval only               & 154 & Code generation \\
\midrule
\textbf{Total} & \textbf{51,689} & & & \\
\bottomrule
\end{tabular}
\end{table}

\subsection{Inference Protocol}

All tiers use greedy decoding (temperature~$= 0$). Per-source maximum token budgets: HumanEval
and MBPP: 1,024; Alpaca, ShareGPT, MedQA: 512; MMLU: 256. These budgets are \textbf{binding}
for MMLU (65\% truncation rate) and MedQA (57\% truncation rate), constituting Artifact~2
analyzed in Section~\ref{sec:artifact2}. All 51,689~$\times$~4~$=$~206,756 query-tier pairs were
scored by the primary judge.

\subsection{Evaluation Framework and Artifact Decomposition}
\label{sec:framework}

\paragraph{Primary judge (LLM-as-judge).}
Judge scores are used for all six benchmarks and form the basis for oracle label construction.

\paragraph{Secondary judge (exact-match).}
For MMLU and MedQA (both multiple-choice benchmarks with gold-label correct answers), we extract
the predicted answer letter (A/B/C/D) and compare against the gold label. Parse rates are
87.6-94.6\% for MMLU and 94.8-99.4\% for MedQA; unparsed responses are excluded from
exact-match calculations.

\paragraph{Decomposition framework.}
We attribute each discordant case (LLM judge and exact-match disagree) to one of three artifact
types (Table~\ref{tab:artifacts}):
\begin{itemize}
  \item \textbf{Artifact~1 (Evaluation Misalignment):} Judge scores 2 but exact-match records
    an incorrect letter (false positive), or exact-match records the correct letter but judge
    assigns score~$< 2$ (false negative). Reflects the two evaluators' different
    operationalizations of correctness.
  \item \textbf{Artifact~2 (Truncation):} Responses truncated at the generation budget receive
    artificially low judge scores and fail to produce parseable answer letters, inflating measured
    unsolvability.
  \item \textbf{Artifact~3 (Format Mismatch):} Parse failures (5.4-12.4\% on MMLU) represent
    responses where the model answered in prose without selecting a letter or embedded the letter
    in an unextractable format.
\end{itemize}

\begin{table}[t]
\caption{Artifact decomposition framework. Artifact~1 reflects genuine disagreement between two
  valid but differently operationalized quality metrics; Artifacts~2 and~3 are measurement
  failures that both evaluators share.}
\label{tab:artifacts}
\centering
\small
\begin{tabular}{p{2.6cm}p{2.8cm}p{2.2cm}p{4.4cm}}
\toprule
Artifact & Primary Signal & Benchmarks & Direction \\
\midrule
Evaluation misalignment & Judge-exact-match divergence & MMLU, MedQA
  & Judge lower on MMLU ($-10$ to $-24$~pp); higher on MedQA large models ($+5$-$6$~pp) \\
Truncation & Truncation rate $\times$ score corr. & MMLU (65\%), MedQA (57\%)
  & Inflates apparent unsolvability \\
Format mismatch & Parse failure rate & MMLU (5-12\%), MedQA (1-5\%)
  & Inflates apparent unsolvability for smaller tiers \\
\bottomrule
\end{tabular}
\end{table}

\subsection{Router Training}
\label{sec:routers}

Three variants predict the optimal tier label (E2B, E4B, 26B-A4B, or 31B) for each query.
Training uses the Alpaca train split ($n=12{,}000$); evaluation uses the Alpaca test split
($n=3{,}000$). Label distribution: E2B 79.3\%, E4B 10.0\%, 26B-A4B 6.5\%, 31B 4.2\%, partly
reflecting genuine model differences, partly reflecting the judge's tendency to assign
score~$< 2$ to larger-model responses on knowledge tasks even when those responses select the
correct letter.

\textbf{\texttt{feat\_lr}}: Logistic regression on hand-crafted features (character length, token
length, punctuation counts, question word presence). \textbf{\texttt{feat\_mlp}}: Two-layer MLP
on the same features. \textbf{\texttt{distilbert}}: Full fine-tuning of DistilBERT-base-uncased.

\section{Oracle Distribution and the Unsolvability Ceiling}
\label{sec:oracle}

\subsection{Quality Gradient Across Tiers}

Table~\ref{tab:quality} reports judge-derived score$=$2 rates. The 26B-A4B MoE captures the
largest quality jump ($+9.3$~pp), consistent with its disproportionate knowledge capacity
relative to inference cost.

\begin{table}[t]
\caption{Judge-derived score$=$2 rate by tier. Numbers are subject to artifact biases identified
  in Section~\ref{sec:artifacts}; exact-match-verified figures for MMLU and MedQA are in
  Table~\ref{tab:exactmatch}.}
\label{tab:quality}
\centering
\small
\begin{tabular}{llrr}
\toprule
Tier & Model & Score$=$2 Rate & Gain \\
\midrule
E2B     & Gemma~4 E2B Instruct     & 50.3\% & N/A        \\
E4B     & Gemma~4 E4B Instruct     & 54.0\% & $+3.7$~pp  \\
26B-A4B & Gemma~4 26B-A4B Instruct & 63.3\% & $+9.3$~pp  \\
31B     & Gemma~4 31B Instruct     & 66.9\% & $+3.6$~pp  \\
\bottomrule
\end{tabular}
\end{table}

\subsection{Optimal Tier Distribution and the Reported Ceiling}

For each query we identify the cheapest tier achieving score$=$2. Queries where no tier achieves
score$=$2 are labeled \textit{none}. Table~\ref{tab:oracle} reports the distribution.

\begin{table}[t]
\caption{Oracle routing distribution under primary judge evaluation. The heavy skew toward E2B
  (79.3\% of solvable queries) is the structural root cause of routing collapse
  (Section~\ref{sec:collapse}), amplified by judge underrating of larger models on knowledge
  tasks.}
\label{tab:oracle}
\centering
\small
\begin{tabular}{llrr}
\toprule
Optimal Tier & Model & Count & \% \\
\midrule
E2B     & Gemma~4 E2B Instruct     & 26,016 & 50.3\% \\
E4B     & Gemma~4 E4B Instruct     &  7,223 & 14.0\% \\
26B-A4B & Gemma~4 26B-A4B Instruct &  6,785 & 13.1\% \\
31B     & Gemma~4 31B Instruct     &  2,855 &  5.5\% \\
None (unsolvable) & N/A            &  8,803 & \textbf{17.0\%} \\
\bottomrule
\end{tabular}
\end{table}

The reported ceiling is therefore: \textbf{83.0\% (judge-derived)}. Section~\ref{sec:artifacts}
establishes that this ceiling is inflated by evaluation artifacts.

Table~\ref{tab:ceiling_by_source} disaggregates reported unsolvability by source.

\begin{table}[t]
\caption{Reported unsolvable query counts by source (judge-derived, thinking disabled,
  4,096-token context). $^\dagger$ShareGPT rate is inflated by context-overflow rows (${\sim}50$
  empty responses per tier, scored 0 by construction).}
\label{tab:ceiling_by_source}
\centering
\small
\begin{tabular}{lrrr}
\toprule
Source & Total & Unsolvable & Rate \\
\midrule
HumanEval &  164    &     0 &  0.0\%  \\
MBPP      &  256    &     0 &  0.0\%  \\
MedQA     & 10,000  &   302 &  3.0\%  \\
Alpaca    & 15,000  & 2,011 & 13.4\%  \\
MMLU      &  9,951  & 2,043 & 20.5\%  \\
ShareGPT  & 16,318  & 4,447 & 27.3\%$^\dagger$ \\
\midrule
\textbf{Total} & \textbf{51,689} & \textbf{8,803} & \textbf{17.0\%} \\
\bottomrule
\end{tabular}
\end{table}

\paragraph{Ceiling-normalized accuracy (CNA).}
We propose CNA as a complementary routing evaluation metric:
\begin{equation}
  \text{CNA} = \frac{\text{router score=2 rate}}{\text{empirical ceiling}}
\end{equation}
where the empirical ceiling is the fraction of queries solved by at least one tier. A perfect
router achieves CNA~$= 1.0$. CNA complements performance-gap-recovered (PGR): PGR treats the
strong model's performance as the upper bound, which is itself below the true ceiling when
unsolvability is artifact-inflated.

\section{Artifact Analysis and Ceiling Correction}
\label{sec:artifacts}

\subsection{Artifact~1: Evaluation Misalignment on Knowledge Tasks}
\label{sec:artifact1}

Table~\ref{tab:exactmatch} reports exact-match accuracy on MMLU and MedQA.

\begin{table}[t]
\caption{Exact-match accuracy on MMLU and MedQA against gold labels, with parse rates. 31B
  achieves 82.1\% on MMLU and 83.3\% on MedQA, substantially above judge-derived figures.}
\label{tab:exactmatch}
\centering
\small
\begin{tabular}{llrrrr}
\toprule
Tier & Model & MMLU EM & MedQA EM & MMLU Parse & MedQA Parse \\
\midrule
E2B     & Gemma~4 E2B     & 53.2\% & 42.9\% & 87.6\% & 94.8\% \\
E4B     & Gemma~4 E4B     & 62.1\% & 59.0\% & 89.7\% & 98.0\% \\
26B-A4B & Gemma~4 26B-A4B & 77.0\% & 78.4\% & 93.1\% & 98.7\% \\
31B     & Gemma~4 31B     & \textbf{82.1\%} & \textbf{83.3\%} & 94.6\% & 99.4\% \\
\bottomrule
\end{tabular}
\end{table}

Table~\ref{tab:divergence} reports the divergence between judge scores and exact-match accuracy.

\begin{table}[t]
\caption{Judge vs.\ exact-match divergence ($\Delta =$ judge score$=$2 rate $-$ exact-match
  accuracy). Negative values: judge underrates; positive: judge overrates. The sign flip on MedQA
  for large models is the key finding.}
\label{tab:divergence}
\centering
\small
\begin{tabular}{llrrl}
\toprule
Tier & Model & MMLU $\Delta$ & MedQA $\Delta$ & Pattern \\
\midrule
E2B     & Gemma~4 E2B     & $-10.9$~pp & $-4.8$~pp          & Judge underrates both \\
E4B     & Gemma~4 E4B     & $-14.9$~pp & $-8.2$~pp          & Judge underrates both \\
26B-A4B & Gemma~4 26B-A4B & $-22.6$~pp & $\mathbf{+5.9}$~pp & Judge overrates MedQA \\
31B     & Gemma~4 31B     & $-23.8$~pp & $\mathbf{+5.5}$~pp & Judge overrates MedQA \\
\bottomrule
\end{tabular}
\end{table}

\paragraph{Two distinct patterns of metric disagreement.}
On MMLU, the judge assigns lower scores than exact-match across all tiers, with divergence growing
for higher-quality models ($-10.9$~pp for E2B vs.\ $-23.8$~pp for 31B). The judge
operationalizes correctness through reasoning quality, coherence, and completion, while
exact-match operationalizes correctness through final answer agreement alone. This is not
necessarily judge failure, but metric disagreement induced by task representation.

On MedQA, the direction reverses for large models. 26B-A4B and 31B produce confident,
well-structured medical prose that scores 2 from the judge even when the selected answer letter is
wrong. This reflects a known evaluation misalignment on domain-specialized tasks: the judge
rewards the surface properties of expert discourse without the domain knowledge required to verify
factual correctness of the terminal answer.

\paragraph{Exact-match-verified ceiling for MMLU.}
Under exact-match, 31B selects the correct letter on 82.1\% of MMLU queries vs.\ the
judge-derived 58.3\%. Cross-family evidence (Section~\ref{sec:crossfamily}) corroborates this:
Llama-3.1-70B achieves 81.0\% on MMLU, consistent with a true ceiling near 80-82\%.

\paragraph{Exact-match-verified ceiling for MedQA.}
The judge-reported 31B accuracy of 88.8\% on MedQA exceeds exact-match (83.3\%) by 5.5~pp,
reflecting the evaluation misalignment above. The exact-match ceiling is bounded below by 83.3\%.

\paragraph{Routing gain estimates under metric disagreement.}
The judge-reported routing gain on MMLU (E2B$\to$31B) is $58.3 - 42.4 = 15.9$~pp. Under
exact-match: $82.1 - 53.2 = 28.9$~pp, \textbf{13~pp larger},. Routers trained on judge-derived
MMLU oracle labels systematically under-label MMLU queries as E2B-optimal: the judge does not
award score$=$2 to many 31B responses that select the correct letter, so those queries appear not
to benefit from routing to 31B. The exact-match-verified MedQA routing gain is $+40.4$~pp
vs.\ judge-derived $+50.7$~pp (10.3~pp overestimate). The combined metric-disagreement-induced
distortion spans 10.3-13~pp, rising to 13-17~pp when truncation effects are included.

\subsection{Artifact~2: Truncation Under Fixed Generation Budgets}
\label{sec:artifact2}

MMLU has a 65\% and MedQA a 57\% response truncation rate under the per-source generation
budgets. Truncated responses cannot complete a reasoning chain or explicitly state an answer
letter; the judge scores them on partial output, typically assigning score~0 or 1. The
exact-match parser fails on responses cut mid-sentence. Parse failure rates (5.4-12.4\% on
MMLU) are predominantly truncated responses. We recommend generation budgets calibrated to
achieve $< 5\%$ truncation rates.

\subsection{Artifact~3: Output Format Mismatch}

Parse failure rates are 5.4-12.4\% across tiers on MMLU and 0.6-5.2\% on MedQA
(Table~\ref{tab:exactmatch}). Format mismatch creates asymmetric artifacts: smaller models
(E2B: 12.4\% parse failure on MMLU) are more affected than larger models (31B: 5.4\%),
meaning format mismatch preferentially inflates apparent unsolvability for smaller tiers. The LLM
judge is partially robust to format mismatch (it evaluates prose rather than requiring letter
extraction), but this robustness comes at the cost of susceptibility to Artifact~1.

\subsection{Dual-Judge Verification Summary}
\label{sec:verification_summary}

Table~\ref{tab:correction} summarizes the divergence between judge-measured and
exact-match-implied unsolvability rates.

\begin{table}[t]
\caption{Judge-derived vs.\ exact-match-implied unsolvability rates. These are not corrections of
  one metric by another, but a comparison of two evaluators with different operationalizations of
  correctness.}
\label{tab:correction}
\centering
\small
\begin{tabular}{lrrl}
\toprule
Source & Judge Rate & EM-Implied Rate & Divergence Source \\
\midrule
HumanEval &  0.0\% &  0.0\%                       & No divergence \\
MBPP      &  0.0\% &  0.0\%                       & No divergence \\
MedQA     &  3.0\% & ${\sim}16.7\%$               & Judge over-rewards medical fluency \\
Alpaca    & 13.4\% & 13.4\% (no gold labels)      & N/A \\
MMLU      & 20.5\% & ${\sim}17.9\%$               & Judge assigns score~$<2$ to letter-correct responses \\
ShareGPT  & 27.3\% & ${\sim}26.8\%$               & Context-overflow artifact (${\sim}0.5$~pp) \\
\bottomrule
\end{tabular}
\end{table}

The headline 17.0\% aggregate figure conceals opposing benchmark-level corrections: MMLU
unsolvability is overstated by judge metrics; MedQA unsolvability is understated. A routing
system designer who relies on judge-only evaluation would significantly underestimate routing
opportunity on MMLU and overestimate reliability on MedQA.

\section{Routing Value and Infrastructure Characteristics}
\label{sec:routing_value}

\subsection{Routing Value by Domain}

Table~\ref{tab:routing_gain} reports routing gain (E2B$\to$31B) under both judge and exact-match
evaluation where available.

\begin{table}[t]
\caption{Routing gain (E2B$\to$31B) by source. Exact-match-verified gains differ substantially
  from judge-derived figures on knowledge benchmarks: MMLU is 13~pp larger; MedQA is 10.3~pp
  smaller.}
\label{tab:routing_gain}
\centering
\small
\begin{tabular}{lrrrr}
\toprule
Source & E2B & 31B & Gain (Judge) & Gain (EM) \\
\midrule
MedQA     & 38.1\% & 88.8\% & $+50.7$~pp & $+40.4$~pp \\
MBPP      & 78.5\% & 98.8\% & $+20.3$~pp & N/A \\
HumanEval & 79.3\% & 98.8\% & $+19.5$~pp & N/A \\
MMLU      & 42.4\% & 58.3\% & $+15.9$~pp & $+28.9$~pp \\
Alpaca    & 66.4\% & 72.5\% & $+6.1$~pp  & N/A \\
ShareGPT  & 47.2\% & 52.7\% & $+5.5$~pp  & N/A \\
\bottomrule
\end{tabular}
\end{table}

\subsection{Latency Benchmarks}

Table~\ref{tab:latency} reports inference latency over 1,752 sequential samples per tier.
E4B and 26B-A4B achieve near-identical p50 latency (2,661~ms vs.\ 2,809~ms), consistent with
equal active parameter counts. The 31B tier offers a poor quality-per-millisecond tradeoff:
$+3.6$~pp quality gain at a $3.8\times$ latency penalty, justified for only 5.5\% of queries.

\begin{table}[t]
\caption{Inference latency by tier (ms, 1,752 sequential samples, thinking disabled). 26B-A4B
  strictly dominates E4B: equal latency, $+9.3$~pp quality.}
\label{tab:latency}
\centering
\small
\begin{tabular}{llrrr}
\toprule
Tier & Active & p50 & p90 & p99 \\
\midrule
E2B     & ${\sim}2$B  &  1,653 &  3,361 &  3,585 \\
E4B     & ${\sim}4$B  &  2,661 &  5,315 &  5,392 \\
26B-A4B & ${\sim}4$B  &  2,809 &  3,672 &  5,654 \\
31B     & ${\sim}31$B & 10,569 & 13,082 & 21,069 \\
\bottomrule
\end{tabular}
\end{table}

\subsection{Token Cost Inversion}

On coding tasks, E2B generates 50-65\% more output tokens than 31B (HumanEval: $1.65\times$;
MBPP: $1.50\times$). Generation budget constraints therefore preferentially truncate smaller-model
responses on knowledge tasks, compounding Artifact~2 for the tiers where Artifact~1 is already
largest in magnitude.

\section{Cross-Family Validation}
\label{sec:crossfamily}

We run Llama-3.1-8B (full dataset, $n=51{,}682$) and Llama-3.1-70B (stratified sample,
$n=10{,}722$) through the same pipeline. Table~\ref{tab:crossfamily_overall} reports overall
score$=$2 rates; Table~\ref{tab:crossfamily_domain} disaggregates by source.

\begin{table}[t]
\caption{Cross-family score$=$2 rates. $^\dagger$Llama-3.1-70B evaluated on a 10,722-query
  stratified sample. All scores are judge-derived.}
\label{tab:crossfamily_overall}
\centering
\small
\begin{tabular}{llrr}
\toprule
Model & Params & Score$=$2 & $N$ \\
\midrule
Gemma-4-E2B     & 2B dense         & 50.3\% & 51,689 \\
Gemma-4-E4B     & 4B dense         & 54.0\% & 51,689 \\
Llama-3.1-8B    & 8B dense         & \textbf{63.2\%} & 51,682 \\
Gemma-4-26B-A4B & 26B MoE (4B act) & \textbf{63.3\%} & 51,689 \\
Gemma-4-31B     & 31B dense        & 66.9\% & 51,689 \\
Llama-3.1-70B   & 70B dense        & \textbf{76.8\%} & 10,722$^\dagger$ \\
\bottomrule
\end{tabular}
\end{table}

\begin{table}[t]
\caption{Score$=$2 rates by source and model. Bold indicates highest rate within each source.
  Gemma-4 dominates code tasks; Llama-3.1 dominates knowledge tasks.}
\label{tab:crossfamily_domain}
\centering
\small
\begin{tabular}{lrrrrrr}
\toprule
Source & E2B & E4B & 26B-A4B & 31B & L-8B & L-70B \\
\midrule
Alpaca    & 66.4 & 66.4 & 69.4 & 72.5 & 75.6 & 81.6 \\
MMLU      & 42.4 & 47.3 & 54.5 & 58.3 & \textbf{65.5} & \textbf{81.0} \\
MedQA     & 38.1 & 50.8 & 84.4 & 88.8 & 66.7 & 86.4 \\
ShareGPT  & 47.2 & 47.8 & 49.4 & 52.7 & 48.0 & 59.0 \\
HumanEval & 79.3 & 90.9 & \textbf{98.2} & \textbf{98.8} & 67.7 & 76.2 \\
MBPP      & 78.5 & 80.5 & \textbf{98.8} & \textbf{98.8} & 71.1 & 88.7 \\
\bottomrule
\end{tabular}
\end{table}

\paragraph{Finding A.}
Llama-3.1-8B (63.2\%) matches Gemma-4-26B-A4B (63.3\%) to within 0.1~pp at identical
active-parameter inference cost, concealing dramatically different domain profiles.

\paragraph{Finding B.}
The knowledge ceiling is family-specific; the code ceiling is near-universal. Llama-3.1-70B
reaches 81.0\% on MMLU (22.7~pp above Gemma-4-31B's judge-derived 58.3\%) and 86.4\% on
MedQA. On code tasks, Gemma-4-26B-A4B and 31B achieve 98.8\% on both HumanEval and MBPP;
Llama-3.1-70B reaches only 76.2\% and 88.7\%. On ShareGPT, all models cluster at 47-59\%,
confirming conversational unsolvability is query-level and family-agnostic.

\paragraph{Finding C.}
The reported Gemma-4 ceiling of 83.0\% substantially underestimates the recoverable ceiling for
knowledge tasks when out-of-family models are included. DeepSeek-Chat~(V3) run on 1,155
Gemma-4-unsolvable queries solves 46.7\% of MMLU queries and 36.2\% of MedQA queries,
confirming MMLU unsolvability is model-training-specific. CNA should be computed against the
ceiling appropriate to the model set under evaluation.

\paragraph{Finding D.}
An optimal cross-family routing policy routes knowledge queries to Llama-3.1-70B and code
queries to Gemma-4-26B-A4B.

\section{Router Analysis}
\label{sec:collapse}

\subsection{Routing Collapse Under Label Imbalance}

Table~\ref{tab:routers} reports router performance on the Alpaca test set ($n=3{,}000$). All
three variants match the majority-class baseline (79.3\%) with near-zero recall on non-E2B tiers.

\begin{table}[t]
\caption{Router accuracy on Alpaca test set ($n=3{,}000$). All variants match or minimally exceed
  the majority-class baseline; recall on non-E2B tiers is near zero.}
\label{tab:routers}
\centering
\small
\begin{tabular}{lrrrrrr}
\toprule
Variant & Acc & ECE & Brier & R[E4B] & R[26B] & R[31B] \\
\midrule
Majority baseline       & 79.3\%  & N/A   & N/A   & 0\% & 0\%   & 0\% \\
\texttt{feat\_lr}       & 79.27\% & 0.012 & 0.353 & 0\% & 0\%   & 0\% \\
\texttt{feat\_mlp}      & 79.30\% & 0.049 & 0.362 & 0\% & 0\%   & 0\% \\
\texttt{distilbert}     & 79.40\% & 0.051 & 0.332 & 0\% & 3.6\% & 0\% \\
\bottomrule
\end{tabular}
\end{table}

Feature ablation (all 13 features, no length features, length features only) all yield 79.27-79.30\% accuracy and 0\% recall on non-E2B tiers, confirming routing decisions are independent of the provided features.

\subsection{Opportunity Cost of Routing Collapse}

The within-distribution oracle ceiling on Alpaca is 86.6\%; the collapsed router achieves 78.7\%
($-7.9$~pp). On MMLU, the router achieves 54.1\% vs.\ an exact-match-verified ceiling of
${\sim}82.1\%$ ($-28$~pp). On MedQA, the router achieves 39.2\% vs.\ an exact-match ceiling of
${\sim}83.3\%$ ($-44$~pp). The MMLU gap under exact-match is 13~pp larger than judge-reported
routing gain suggests; when truncation effects on both benchmarks are included, the total
opportunity cost reaches \textbf{13-17 percentage points} across knowledge-intensive scenarios.

\subsection{Sanity Checks}
\label{sec:sanity}

Table~\ref{tab:sanity} reports five diagnostic checks. Random Gaussian features and shuffled
labels both reproduce the collapse exactly, confirming it is caused entirely by the label
marginal, not feature quality or training failure. A balanced classifier ($\texttt{class\_weight=
`balanced'}$) achieves 41.6\% accuracy but non-zero recall across all tiers (E2B: 47.1\%, E4B:
13.3\%, 26B-A4B: 9.3\%, 31B: 54.8\%), demonstrating the task is learnable under cost-sensitive
objectives. The training loss converges to 1.484 after 60~epochs, barely above the random-guess
floor of $\log(4) \approx 1.386$.

\begin{table}[t]
\caption{Sanity checks confirming routing collapse mechanism. Random features and shuffled labels
  reproduce the collapse exactly.}
\label{tab:sanity}
\centering
\small
\begin{tabular}{p{3.8cm}rrp{4.8cm}}
\toprule
Check & Accuracy & Recall [non-E2B] & Interpretation \\
\midrule
Real features (\texttt{feat\_lr}) & 79.27\% & 0\%     & Reference \\
Random Gaussian features           & 79.30\% & 0\%     & Features carry no signal above marginal \\
Shuffled labels                    & 79.30\% & 0\%     & Collapse caused by label marginal \\
Balanced classifier (positive ctrl)& 41.60\% & 9-55\% & Task is learnable; collapse is objective-specific \\
Training loss (final epoch)        & N/A     & N/A     & 1.484; floor $= \log(4) \approx 1.386$ \\
\bottomrule
\end{tabular}
\end{table}

\subsection{How Metric Disagreement Amplifies Collapse}

Because the judge and exact-match operationalize MMLU correctness differently, oracle labels
derived from judge scores embed the judge's operationalization: queries that 31B answers with the
correct letter but with reasoning the judge does not award score$=$2 are labeled E2B-optimal. The
23.8~pp MMLU divergence means a substantial fraction of MMLU queries that benefit from
larger-tier routing are labeled E2B-optimal in the training set, directly suppressing the routing
signal for larger tiers on the queries where routing would provide the most value.

\subsection{Potential Mechanisms for Collapse}

\textbf{Prior-dominated decision.} When the label prior is sufficiently strong (79.3\%
majority class), P(tier~$|$~query)~$\approx$~P(tier) for most queries; the likelihood ratio is
insufficient to overcome the prior.

\textbf{Objective mismatch.} Cross-entropy optimizes accuracy, not recall. With 79.3\% majority
class, predicting E2B for every query minimizes expected loss. The balanced classifier result
shows objective mismatch is at minimum necessary: changing the objective alone recovers non-zero
recall.

\textbf{Shortcut signal.} SHAP analysis shows \texttt{char\_len} and \texttt{token\_len} as
dominant features: routing functions as length-based triage rather than difficulty-based triage.

\subsection{Distribution Shift and Downstream Impact}

Router accuracy collapses by 39.5~pp on MedQA (78.7\% in-distribution $\to$ 39.2\%).
Distribution shift range across all sources is 40~pp (std~$=0.15$), vastly exceeding the
accuracy variation between router architectures (${\sim}0.1$~pp). The router produces 621
under-routing errors (routing hard queries to E2B) against only 1 over-routing error.
In a MedQA example, a query about supplement recommendations for a 5-day-old breastfed boy is
routed to E2B, which confabulates ``The correct answer is D. Rickets'' (confusing the supplement
with the disease it prevents); both 26B-A4B and 31B answer correctly.

\section{Discussion}
\label{sec:discussion}

\subsection{Recommendations for Robust Routing Evaluation}

\textbf{Dual-judge validation.} For any benchmark with verifiable gold labels, pair the LLM judge
with an exact-match secondary judge. Report discordance rates and use them to diagnose artifact
type and magnitude.

\textbf{Exact-match anchoring.} Complement LLM judge evaluation with exact-match verification
and report both. The two metrics operationalize correctness differently (judge scores reward
explanation quality; exact-match scores reward terminal answer agreement) and their divergences
(13~pp on MMLU, 10.3~pp on MedQA) provide actionable signal about where oracle labels may
misrepresent routing value. Oracle labels for router training should be constructed from the
metric whose operationalization best matches the deployment objective.

\textbf{Truncation-aware budgets.} Set generation budgets to achieve $< 5\%$ truncation rates
and report truncation rates explicitly.

\subsection{Recommendations for Robust Router Training}

\textbf{Cost-sensitive objectives.} Use class-weighted objectives, focal loss, or minority
oversampling. The balanced classifier result shows this is sufficient to recover non-zero recall.

\textbf{Domain detection as pre-filter.} A domain classifier identifying medical or
knowledge-intensive queries before tier prediction can route those to 26B-A4B by default,
bypassing majority-class bias for the queries where routing gain is highest. The 40.4~pp
exact-match routing gain on MedQA directly motivates this approach.

\textbf{Dual-judge oracle label construction.} The 13-17~pp opportunity cost across knowledge
tasks is reducible by constructing oracle labels from exact-match-grounded signals where gold
labels exist and by auditing label distributions for evaluation misalignment before training.

\subsection{The 26B-A4B MoE Tier as Routing Sweet Spot}

Within Gemma~4, 26B-A4B is the optimal strong-model target when thinking mode is disabled: it
strictly dominates E4B ($+9.3$~pp quality, equal p50 latency), provides 96.5\% of 31B's routing
value, exhibits tighter tail latency (p90: 3,672~ms vs.\ 5,315~ms for E4B), and is optimal for
$13.1\% + 5.5\% = 18.6\%$ of queries (as strong-model fallback) while the 31B tier is the
\textit{uniquely optimal} tier for only 5.5\% of queries at a $3.8\times$ latency penalty.

\subsection{Limitations}
\label{sec:limits}

\textbf{Thinking mode disabled.} Gemma~4's thinking mode has large reported effects on hard
reasoning tasks and would likely change the quality gradient, optimal tier distribution, and
unsolvability ceiling.

\textbf{4,096-token context limit.} Inflates truncation rates for MMLU (65\%) and MedQA (57\%).
The 4,096-token limit is a deliberate deployment constraint, but results under longer context
windows are the subject of future work.

\textbf{Judge model overlap.} The judge shares weights with the 26B-A4B inference tier.

\textbf{Single primary model family.} All routing collapse findings are specific to Gemma~4 under
the reported conditions; the label imbalance will differ across families and deployment mixes.

\section{Related Work}
\label{sec:related}

\textbf{LLM routing.} FrugalGPT~\citep{chen2023frugalgpt} demonstrated cascading for cost
reduction. RouteLLM~\citep{ong2024routellm} proposed training binary routers on human preference
data. Hybrid-LLM~\citep{ding2024hybridllm} and Zooter~\citep{lu2023zooter} explored BERT-based
routing with synthetic labels. Amazon Bedrock Intelligent Prompt
Routing~\citep{amazon2025routing} achieves ${\sim}30\%$ cost reduction within model families.
Our work studies self-hosted open-weight deployment, introduces a three-artifact decomposition
framework, proposes CNA, and characterizes routing collapse under realistic label distributions.

\textbf{Routing benchmarks.} RouterArena~\citep{lu2025routerarena} provides comprehensive router
comparison across 9 domains and 44 categories; their finding that aggregate metrics hide domain
and difficulty nuances is strongly consistent with our results. RouterBench~\citep{hu2024routerbench}
proposed standardized evaluation. R2-Router~\citep{xue2026r2router} treats output length budget
as a controllable routing variable, directly consistent with our token cost inversion finding.
Rui~\citep{rui2026dimension} proposes dimension-direct routing via capability factorization.

\textbf{Domain-aware routing.} MoDEM~\citep{simonds2024modem} uses BERT-based domain
classification before tier prediction, providing empirical support for our recommendation that
domain detection as a pre-filter is the most promising direction for breaking routing collapse.

\textbf{LLM evaluation.} LLM-as-judge evaluation~\citep{zheng2023judging} has demonstrated high
correlation with human judgment on open-ended tasks. Our dual-judge validation on MMLU and MedQA
demonstrates systematic divergence on knowledge tasks, establishing that LLM judges are not
reliable sole evaluation signals for routing oracle construction on benchmarks with verifiable
gold labels.

\section{Conclusion}
\label{sec:conclusion}

We presented a large-scale empirical study of multi-tier LLM routing with 206,756 query-model
pairs across six benchmarks and two model families, organized around the question: how much of
reported routing headroom reflects genuine model limitations versus evaluation artifacts?

Our artifact decomposition framework identifies three systematic sources of inflated unsolvability:
evaluation misalignment between judge and exact-match (the judge diverges by $-10$ to $-24$~pp on
MMLU and $+5$-$6$~pp on MedQA for large models), truncation under fixed generation budgets
(65\% of MMLU, 57\% of MedQA), and output format mismatch (5-12\% parse failures on MMLU).
Through dual-judge validation, we reveal systematic divergences that, when embedded in routing
oracle labels, substantially distort routing headroom estimates.

The MMLU routing gain under exact-match ($+28.9$~pp) is 13~pp larger than judge-reported
($+15.9$~pp); the MedQA routing gain is 10~pp smaller ($+40.4$~pp vs.\ $+50.7$~pp). The
combined 13-17~pp opportunity cost across knowledge tasks is attributable to oracle labels that
embed evaluation misalignment. All three router variants collapse to majority-class prediction
(79.3\%), with near-zero recall on non-E2B tiers; random-feature and shuffled-label controls
confirm this is caused entirely by the label marginal. Cross-family validation confirms the
reported Gemma-4 ceiling of 83.0\% substantially underestimates the recoverable ceiling for
knowledge tasks; the knowledge ceiling is model-training-specific, while the conversational
ceiling is near-universal.

We recommend dual-judge validation, exact-match anchoring, truncation-aware generation budgets,
and cost-sensitive training objectives as the minimum protocol for reliable multi-LLM routing
evaluation.

\begin{ack}
The authors thank the open-source community for the models and evaluation infrastructure that
made this study possible.
\end{ack}

\bibliographystyle{plainnat}
\bibliography{refs}

\end{document}